\documentclass{article}
\usepackage[utf8]{inputenc}
\DeclareUnicodeCharacter{2212}{\textminus}

\usepackage{arxiv}

\usepackage[utf8]{inputenc} 
\usepackage[T1]{fontenc}    
\usepackage{hyperref}       
\usepackage{url}            
\usepackage{booktabs}       
\usepackage{amsfonts}       
\usepackage{nicefrac}       
\usepackage{microtype}      
\usepackage{lipsum}		
\usepackage{graphicx}
\usepackage{doi}
\usepackage[square, numbers]{natbib}

\title{ChatGPT for automated grading of short answer questions in mechanical ventilation}

\author{ \href{https://orcid.org/0009-0005-7946-5154}{\includegraphics[scale=0.06]{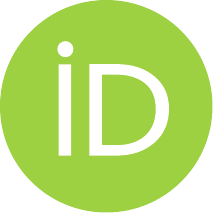}\hspace{1mm}Tejas Jade} \\
        Shri Atal Bihari Vajpayee Medical College and Research Institute\\
	Bowring and Lady Curzon Hospitals\\
	Rajiv Gandhi University of Health Sciences\\
	Bengaluru, 56001\\
	\texttt{tejasjade03@gmail.com} \\
	\And
	\href{https://orcid.org/0000-0003-2901-496X}{\includegraphics[scale=0.06]{orcid.pdf}\hspace{1mm}Alex Yartsev} \\
	MBBS, BappSci(MRS), MClinEd, FCICM\\
	Faculty of Medicine and Health, Sydney Medical School\\
	The University of Sydney \\
        Intensive Care Service, Westmead Hospital\\
	\texttt{alex.yartsev@sydney.edu.au} \\
}

\renewcommand{\shorttitle}{ChatGPT for Automated grading of SAQs}

\hypersetup{
pdftitle={A template for the arxiv style},
pdfsubject={q-bio.NC, q-bio.QM},
pdfauthor={David S.~Hippocampus, Elias D.~Striatum},
pdfkeywords={First keyword, Second keyword, More},
}

\begin{document}
\maketitle

\begin{abstract}
    Standardised tests using short answer questions (SAQs) are common in postgraduate education.  Large language models (LLMs) simulate conversational language and interpret unstructured free-text responses in ways aligning with applying SAQ grading rubrics, making them attractive for automated grading. We evaluated ChatGPT 4o to grade SAQs in a postgraduate medical setting using data from 215 students (557 short-answer responses) enrolled in an online course on mechanical ventilation (2020–2024). Deidentified responses to three case-based scenarios were presented to ChatGPT with a standardised grading prompt and rubric. Outputs were analysed using mixed-effects modelling, variance component analysis, intraclass correlation coefficients (ICCs), Cohen’s kappa, Kendall’s W, and Bland-Altman statistics. ChatGPT awarded systematically lower marks than human graders with a mean difference (bias) of −1.34 on a 10-point scale. ICC values indicated poor agreement individual-level agreement (ICC1=0.086), and Cohen’s kappa (−0.0786) suggested no meaningful agreement. Variance component analysis showed minimal variability among the five ChatGPT sessions (G-value = 0.87), indicating internal consistency but divergence from the human grader. The poorest agreement was observed for evaluative and analytic items, whereas checklist and prescriptive rubric items had less disagreement. We caution against the use of LLMs in grading postgraduate coursework. Over 60\% of ChatGPT-assigned grades differed from human grades by more than acceptable boundaries for high-stakes assessments. 
\end{abstract}

\section{Introduction}
The use of publicly available large language models to automate the
grading of
essays \cite{kasneciChatGPTGoodOpportunities2023, khanChatGPTReshapingMedical2023}
is an attractive response to the increasing demands on the time of the
faculty. Grading assignments can be resource-intensive with turnaround
times ranging from a few days to a few weeks. Essay questions and short
answer questions (SAQs) are written in natural language, often test a
mixture of recalled knowledge and analysis, and range between a phrase
and several paragraphs in length, with the focus of assessment being the
content of the answers rather than the writing
style \cite{burrowsErasTrendsAutomatic2015}.
The development of Automated Essay Scoring (AES) and Automated Short
Answer Grading (ASAG) systems is well
documented \cite{burrowsErasTrendsAutomatic2015, brewAutomatedShortAnswer2013}
where AES's focus on language and grammar and ASAGs focus on
content\cite{brewAutomatedShortAnswer2013}.
These proprietary systems have largely remained inaccessible to the
learner and teacher owing to the prohibitive cost of licensing. In
contrast, Large Language models (LLM) like ChatGPT are readily available
to the general public. The use of LLMs as an automated grader has been
described in language courses
on English \cite{mizumotoExploringPotentialUsing2023}
and
Japanese \cite{obataAssessmentChatGPTsValidity2023}
as well as courses on
astronomy \cite{impeyUsingLargeLanguage2024}
and curriculum
development \cite{jackariaComparativeAnalysisRating2024}.

At its core, grading of SAQs relies on natural language
processing \cite{burrowsErasTrendsAutomatic2015a}
and LLMs are excellent at these
tasks \cite{hajikhaniCriticalReviewLarge2024}.
The application of LLM processing to grading essays and SAQ assessments
is a logical application of these capabilities, given the increasing use
of LLMs by learners and educators
\cite{baigChatGPTHigherEducation2024, davisTemperatureFeatureChatGPT2024}.
In medical education, there have been attempts to assess the performance
of ChatGPT in grading standardized
tests \cite{kungPerformanceChatGPTUSMLE2023}
including some in areas of specialization like
ophthalmology \cite{antakiEvaluatingPerformanceChatGPT2023},
pediatrics \cite{leChatGPTYieldsPassing2024}
and
radiology \cite{bhayanaPerformanceChatGPTRadiology2023}.
To our knowledge, there has been only one paper evaluating the efficacy
of ChatGPT in grading undergraduate short
answers \cite{morjariaExaminingEfficacyChatGPT2024}
and none in the field of postgraduate medical education, where content
expertise is necessary for accurate grading. Content-dense fields where
assessment requires professional judgment and analysis may not be
susceptible to accurate LLM-mediated grading when unmodified questions,
answers and rubrics intended for human experts are presented to the LLM
as prompts. To evaluate this assertion, we describe the use of ChatGPT
for grading SAQs in a postgraduate medical education course on the
subject of mechanical ventilation.

\section{Methods}
An observational retrospective audit of existing student answers was
performed. The answers were retrieved from the University of Sydney's
database (Canvas), for a postgraduate online course, the objective of
which is to familiarise students with the practical aspects of managing
mechanically ventilated critically ill patients. The course has a twelve
week duration and presents the students with three summative assessments
which are presented as case-based discussions representative of
real-world scenarios. Since previous papers on the subject have no
existing rationale for sample size estimation, a sample of 557 answers
was retrieved as a sample of convenience, representing the entire
available database (from 5 years, 2020 to 2024). De-identified answers,
question stems and marking rubrics were converted into the Microsoft
Word (.docx) file format and submitted together as prompts to the
ChatGPT 4o API (model ``gpt-4-turbo") with maximum token value of 4000
and a temperature setting of 0. The API was accessed in multiple
separate sessions to evaluate the effects of the inherent probabilistic
variability in LLM responses on the variability of grades between
sessions using G theory calculations. Outputs produced by the LLM were
appended to a comma-separated value (CSV) file and analysed using the R
statistical software (R version 4.4.1). Averaged LLM grades were
compared to the grades recorded by a human content expert using
intraclass correlation coefficients (ICCs), Cohen's kappa, Kendall's W,
and Bland-Altman statistics.

\section{Results}
G theory analysis determined that five LLM input sessions achieved a
standard of inter-rater variability that would be acceptable for experts
grading high-stakes
assessments\cite{peetersEducationalTestingValidity2013a}
($\sigma\textsubscript{rater}\textsuperscript{2}= 0.0004, G-value = 0.93$) with
the variance component analysis summarised in \textbf{\autoref{tab:variance_components}} . Variance
attributable to differences among answers accounted for 71.08\% of the
total variance ($\sigma\textsubscript{answer}\textsuperscript{2}=4.116$).

\begin{table}[htbp]
  \centering
  \caption{Variance Components and Reliability Coefficients}
  \begin{tabular}{l@{\hspace{3em}}c@{\hspace{3em}}c@{\hspace{3em}}c}
    \toprule
    \textbf{Source} & \textbf{var} & \textbf{percent} & \textbf{n} \\
    \midrule
    Question & 4.11 & 71.08\% & 1 \\
    Session & 0.00 & 0.01\% & 1 \\
    Residual & 1.67 & 28.91\% & 1 \\
    \midrule
    Total Variance & 5.79 & 100\% & - \\
    \midrule
    Adjusted Error Variance & 0.34 & 5.78\% & 5 \\
    (5 raters) & & & \\
    \midrule
    G-Value & 0.93 & - & - \\
    \bottomrule
  \end{tabular}
  \vspace{1em}  
  
  \label{tab:variance_components}
\end{table}

\begin{figure}
    \centering
    \caption{Frequency distribution of grades }
    \includegraphics[width=1\linewidth]{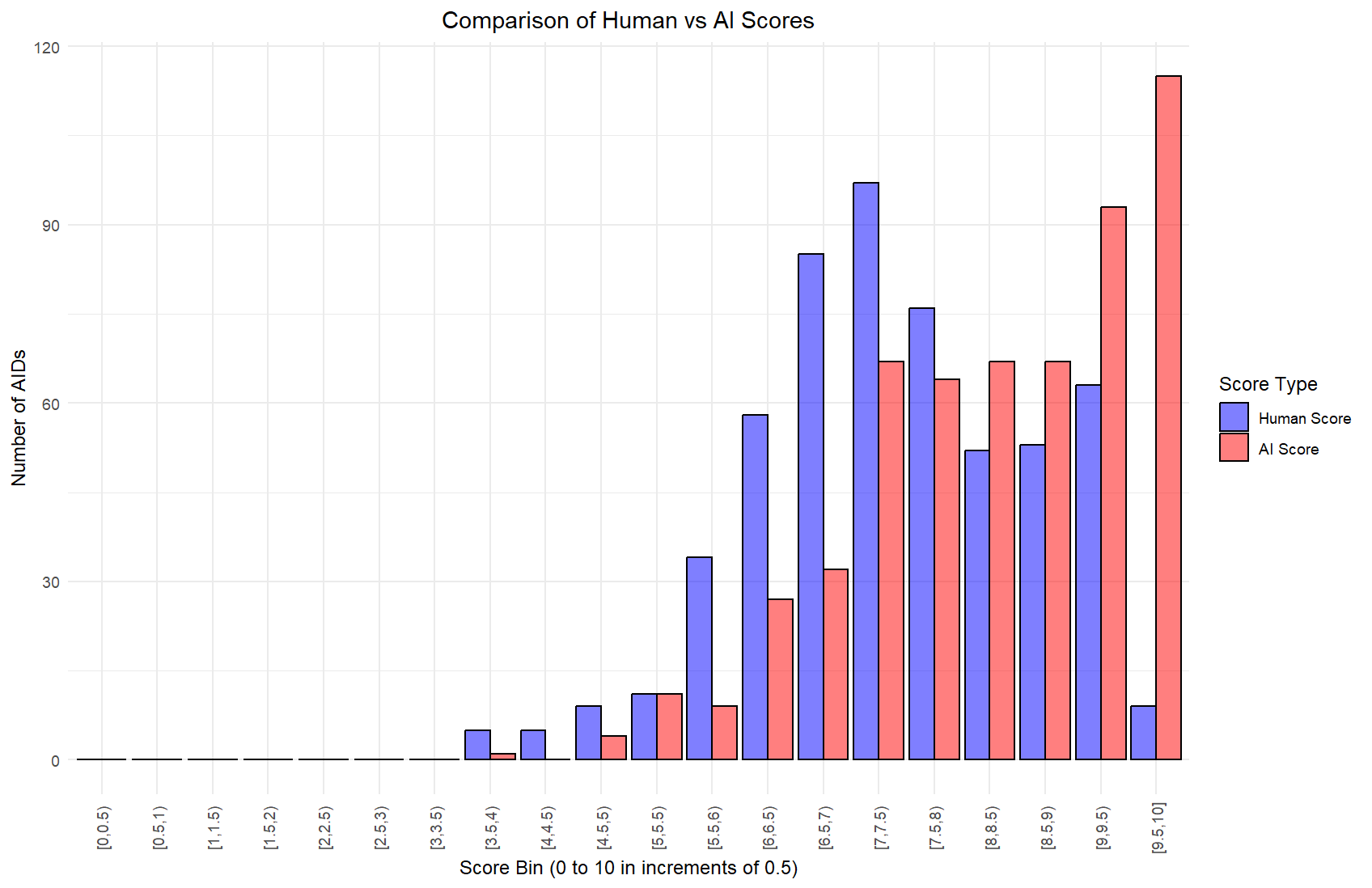}
    
    \label{img:freq_distribution}
\end{figure}

\begin{figure}
    \centering
    \caption{Histogram of grade discrepancy between AI and human graders}
    \includegraphics[width=1\linewidth]{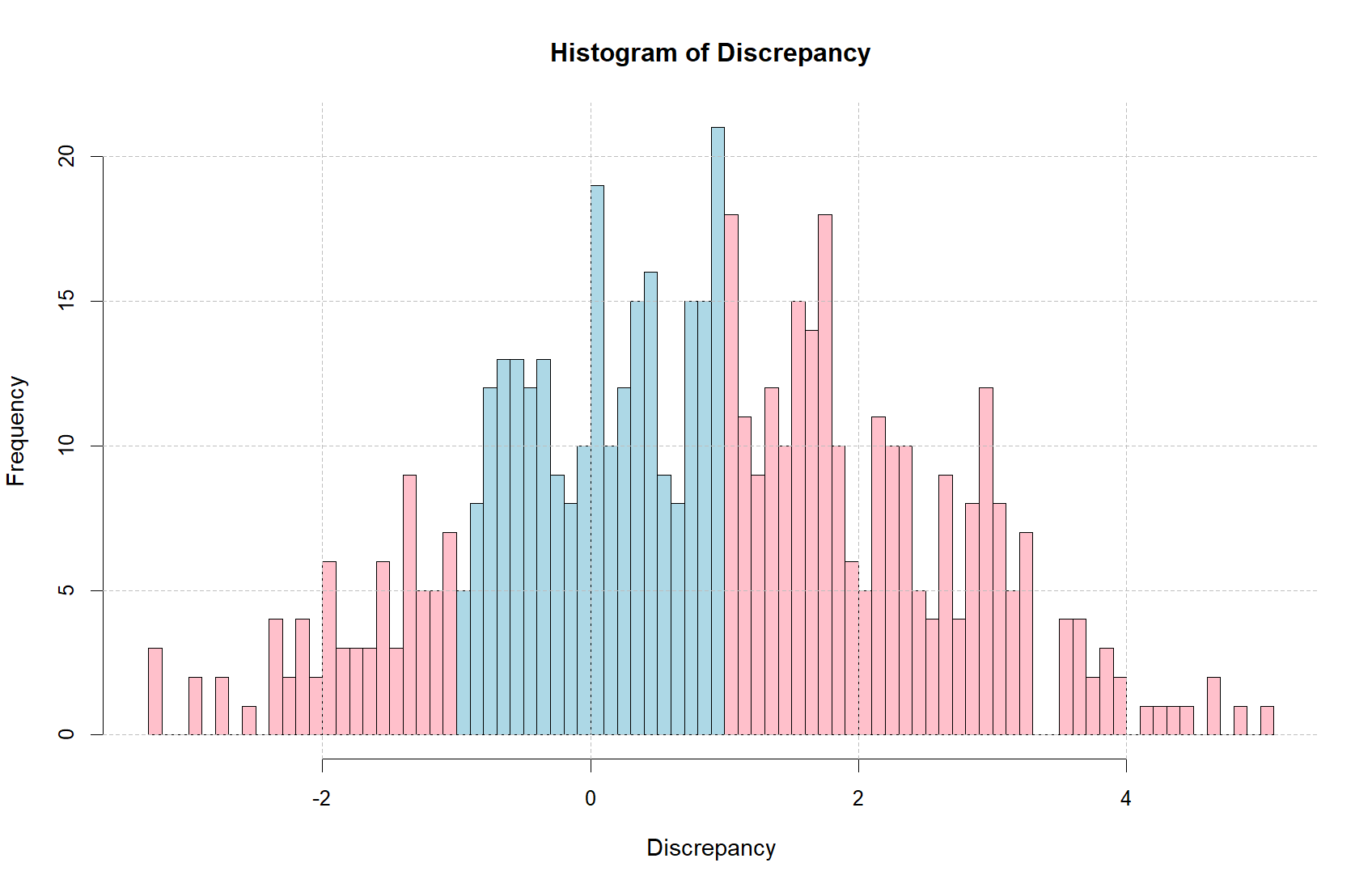}
    
    \label{img:histogram_discrepancy}
\end{figure}

Statistical analysis of agreement between human and LLM is presented in
\textbf{\autoref{tab:stat_analysis}.} Overall, the LLM disagreed with the human grader
significantly, and marked more harshly, being more reluctant to award
full marks (\textbf{\autoref{img:freq_distribution}}). The Intraclass Correlation Coefficients
(ICCs) indicate poor agreement between human and AI scores at the
individual level ($ICC1 = 0.086, ICC2=0.201, ICC3=0.269$). Cohen's kappa
yielded a value of $\kappa = -0.0786, p = 0.0564$ , suggesting no meaningful
agreement between raters. Kendall's W observed a weak positive
association between human scores and average AI scores ($\tau=0.201,
2-sided pvalue = =\textless{} 2.22e-16$). Bland-Altman analysis revealed
that, despite the small proportion of extreme discrepancies (4.49\%),
the mean difference was -1.34 (AI scores were systematically lower than
human scores). The standard deviation of the differences was 1.55 (LoA
from -3.8338 to 2.2368), which falls outside the usual acceptable range
of disagreement between graders in a high-stakes exam where the rating
scale is out of 10 ( +/-
10\%) \cite{peetersEducationalTestingValidity2013a, johnsonAssessingPerformanceDesigning2009}.
\textbf{\autoref{img:histogram_discrepancy} }represents the discrepancies between averaged AI and human
marks as a histogram of frequency, coding the range outside of these
acceptable limits in red. Of the student answers, 63 \% (342) were
graded with an unacceptably high discrepancy between human and AI
raters.

\begin{figure}
    \centering
    \caption{ANOVA of each rubric item}
    \includegraphics[width=1\linewidth]{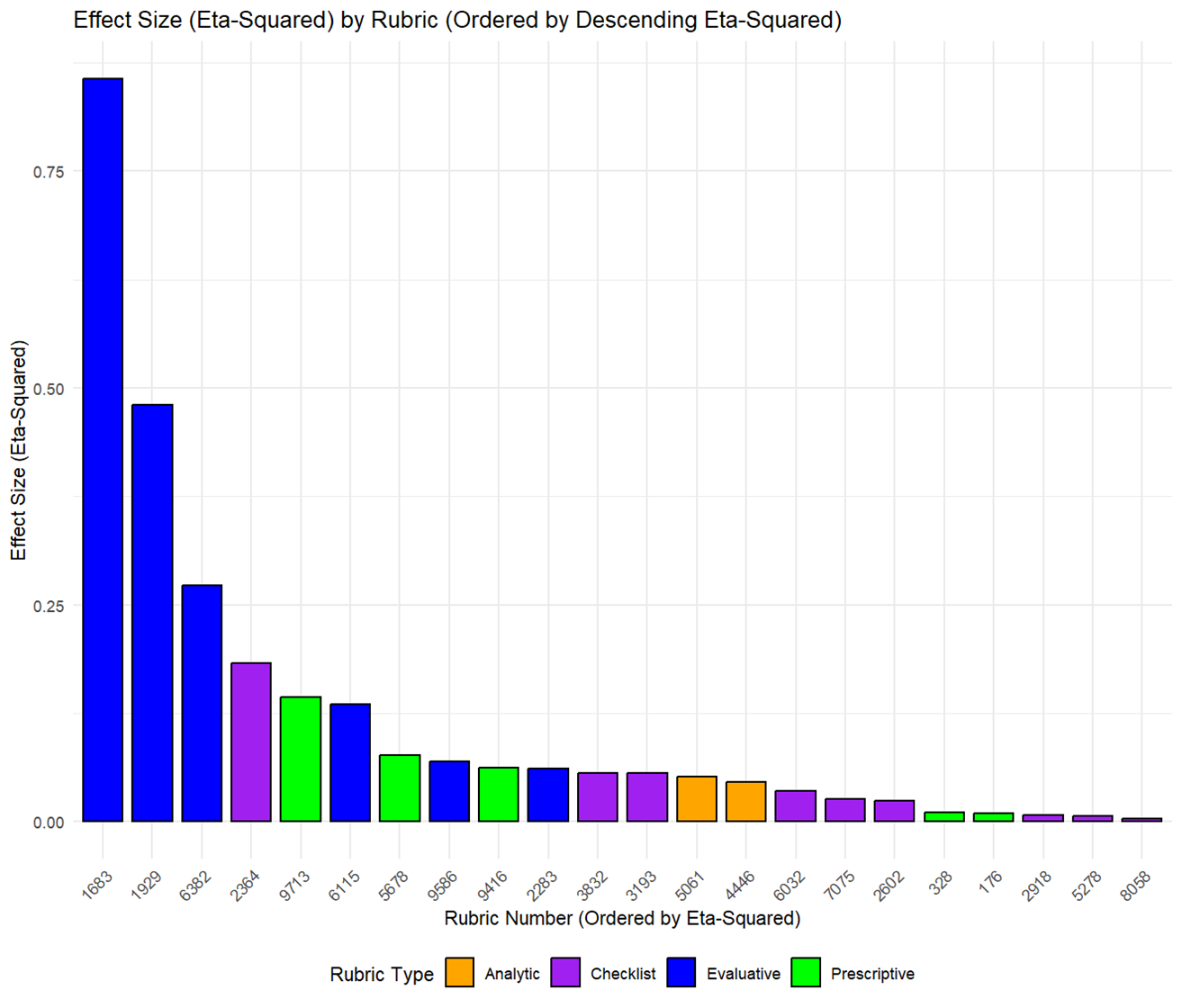}
    
    \label{img:ANOVA_each_item}
\end{figure}

\begin{table}[htbp]
  \centering
  \caption{Statistical analysis of agreement between human and averaged LLM grades}
  \begin{tabular}{l@{\hspace{4em}}c@{\hspace{2em}}c@{\hspace{2em}}c@{\hspace{2em}}c}
    \toprule
    \multicolumn{5}{l}{\textbf{Intraclass Correlation Coefficients (ICC) Results}} \\
    \midrule
    \textbf{Type} & \textbf{ICC} & \textbf{Lower\_Bound} & \textbf{Upper\_Bound} & \textbf{P\_Value} \\
    \midrule
    Single Raters & 0.086 & 0.0031 & 0.17 & 0.021 \\
    Absolute (ICC1) & & & & \\
    \midrule
    Single Random & 0.201 & 0.038 & 0.34 & 0 \\
    Raters (ICC2) & & & & \\
    \midrule
    Single Fixed & 0.269 & 0.1909 & 0.34 & 0 \\
    Raters (ICC3) & & & & \\
    \midrule
    Average Raters & 0.158 & 0.0062 & 0.29 & 0.021 \\
    Absolute (ICC1k) & & & & \\
    \midrule
    Average Random & 0.335 & 0.0732 & 0.51 & 0 \\
    Raters (ICC2k) & & & & \\
    \midrule
    Average Fixed & 0.424 & 0.3207 & 0.51 & 0 \\
    Raters (ICC3k) & & & & \\
    \midrule
    \multicolumn{5}{l}{\textbf{Bland-Altman statistics}} \\
    \midrule
    \textbf{Metric} & \textbf{Mean} & \textbf{Standard Deviation} & \textbf{Upper Limit of} & \textbf{Lower Limit of} \\
     & \textbf{Difference} & \textbf{of Differences} & \textbf{Agreement (Mean +} & \textbf{Agreement (Mean -} \\
     & \textbf{(Bias)} & & \textbf{1.96*SD)} & \textbf{1.96*SD)} \\
    \midrule
    Value & -1.34495 & 1.972839 & 2.521815 & -5.211714 \\
    \midrule
    \multicolumn{5}{l}{\textbf{Cohen's kappa}} \\
    \midrule
    \textbf{Metric} & \textbf{Kappa} & \textbf{p value} & & \\
    \midrule
    Value & -0.0786 & <0.000 & & \\
    \midrule
    \multicolumn{5}{l}{\textbf{Kendall's W}} \\
    \midrule
    \textbf{Metric} & \textbf{T} & \textbf{2-sided p value} & & \\
    \midrule
    Value & 0.201 & <0.000 & & \\
    \bottomrule
    \end{tabular}  
  \label{tab:stat_analysis}
\end{table}

ICC for each rubric item were individually analysed to assess the
contribution of specific wording and structure to poor overall
agreement. The best alignment was observed for checklist and
prescriptive rubric items, whereas evaluative and analytic items had a
greater degree of disagreement (examples of each style of rubric are
offered in \textbf{\autoref{tab:rubric_examples}}; results of ICCs for rubric types in
\textbf{\autoref{tab:icc_scores}})\textbf{.} Analysis of Variance (ANOVA) with effect
size estimation was performed, specifically focusing on eta-squared ($\eta$²)
as a measure of the proportion of variance explained by each rubric
item. These data are visualised as \textbf{\autoref{img:ANOVA_each_item}}. The poorest agreement
between AI and human grading was observed for rubric items with
subjective language or those that depended on a specialised expert
knowledge base.

\begin{table}[htbp]
\vspace{-10pt}
\centering
\caption{Examples of rubrics}
\begin{tabular}{p{6em}p{16em}p{21em}}
\toprule
\textbf{Rubric type} & \textbf{Question stem} & \textbf{Rubric} \\
\midrule
Checklist & ``What are the management options?'' & The following modalities needs to be mentioned as options, 1 mark for each: \\
 & & $\bullet$ Standard oxygen delivery devices \\
 & & $\bullet$ HFNP (high flow nasal prongs) \\
 & & $\bullet$ NIV (non invasive ventilation) \\
 & & $\bullet$ Intubation \\
\midrule
Prescriptive & Elaborate on the option of using NIV (non invasive ventilation) & 1 mark: excellent answer \\
 & & 0.75 marks: strong answer \\
 & & 0.5 mark: satisfactory answer \\
 & & 0.25 mark: minimal answer \\
 & & 0 mark: not evident \\
 & & \\
 & & \textit{Excellent}: correct, well structured and detailed \\
 & & \textit{Strong}: correct and detailed or well structures \\
 & & \textit{Satisfactory}: correct \\
 & & \textit{Minimal}: incorrect, and/or poorly structured, and/or lacking in detail \\
 & & \textit{Not evident}: irrelevant or omitted \\
\midrule
Analytic & If his oxygenation continues to worsen, how will you change your management to rescue the situation? Give at least two rescue techniques as options. & Marks in 3 domains: answer quality, number of options, accuracy of options \\
 & & 2 mark: excellent answer; two valid options \\
 & & 1.5 marks: strong answer; one valid option and another option that was not on the list but which is still reasonable, or a non-ventilator strategy (eg. ECMO) \\
 & & 1.0 mark: satisfactory answer, with one good option (and the other option is unsatisfactory because it does not apply, or would not benefit) \\
 & & 0.5 mark: minimal answer, eg. a misunderstanding of how to ventilate severe refractory ARDS \\
 & & 0 mark: not evident \\
\midrule
Evaluative & What is the rationale for the rescue techniques you have given? How does the evidence support them? & 4 mark: excellent answer, describes HOW the evidence supports the choices of rescue strategies, eg. not necessarily quoting specific mortality statistics, but outlining the role for the strategy and the strength of the recommendations to support it \\
 & & 3 marks: strong answer, describes which choices are supported by the evidence and which are not, but does not have any specific outcomes data \\
 & & 2 marks: satisfactory answer, describes correct choices but has little grasp of the evidence behind them \\
 & & 1 mark: minimal answer, no specific evidence mentioned, or only online lectures or websites are quoted \\
\bottomrule
\end{tabular}
\label{tab:rubric_examples}
\end{table}

\begin{table}[htbp]
\centering
\caption{Average ICC Scores by Rubric Type}
\begin{tabular}{lrrrrrr}
\toprule
\textbf{Type} & \multicolumn{1}{c}{\textbf{ICC1}} & \multicolumn{1}{c}{\textbf{ICC2}} & \multicolumn{1}{c}{\textbf{ICC3}} & \multicolumn{1}{c}{\textbf{ICC1k}} & \multicolumn{1}{c}{\textbf{ICC2k}} & \multicolumn{1}{c}{\textbf{ICC3k}} \\
\midrule
\textbf{Analytic} & -0.013852 & 0.038300 & 0.042043 & -0.030570 & 0.073181 & 0.080059 \\
\textbf{Checklist} & 0.164240 & 0.197975 & 0.214298 & 0.255618 & 0.312441 & 0.333178 \\
\textbf{Evaluative} & -0.254910 & 0.046673 & 0.067620 & -2.203537 & 0.081282 & 0.114124 \\
\textbf{Prescriptive} & 0.020421 & 0.082671 & 0.092168 & 0.026359 & 0.150615 & 0.166609 \\
\bottomrule
\end{tabular}
\label{tab:icc_scores}
\end{table}

\section{Discussion}
Natural language processing by LLMs may produce responses that appear
plausible and similar to human experts on a superficial level, but the
underlying processes that produce these responses are cardinally
different. LLM responses to prompts are generated by modelling the
likelihood of word sequences (``tokens'') that follow the input prompt
sequence, on the basis of patterns extrapolated from training
data \cite{zhaoSurveyLargeLanguage2024}.
A prompt is fitted to a prompt template (identified by parsing the
prompt to recognise key patterns, such as question formats, commands, or
structured input) to convert it into a text generation
problem \cite{schulhoffPromptReportSystematic2024, liuPretrainPromptPredict2021}.
A sequence of words and phrases is then generated, each response
constructed sequentially from a string of tokens by repeated
reassessment of the total response by a neural network
(``transformer'') \cite{vaswaniAttentionAllYou2023}.
The selection of each subsequent token is based on a data set that
contains strings of
text \cite{minaeeLargeLanguageModels2024}
converted, based on their semantics, into a vector in a higher
dimension, with weights given to each dimension
(``embedding'') \cite{camacho-colladosWordSenseEmbeddings2018}.
On the basis of the data set, novel combinations of tokens can be
generated probabilistically to ``predict'' the desired language content
of the response, hence the term ``language model''. Larger more detailed
data sets permit better predictive performance for the model, and
``large'' language models use data sets ranging anywhere from millions
to hundreds of billions of
parameters \cite{naveedComprehensiveOverviewLarge2024}.
A probabilistic randomization of tokens is built into the selection,
which is designed to mimic the unpredictability of conversational
language, and this can be controlled by the user through the adjustment
of a ``temperature''
setting \cite{davisTemperatureFeatureChatGPT2024, BestPracticesPrompt}.
The interpretation of an SAQ response by an expert can be described
along parallel lines. At a fundamental level, the steps that are
involved are
similar \cite{wolfeUncoveringRatersCognitive2005},
consisting of reading the student response, separating it into component
parts according to the rubric, evaluating each response by comparing it
to the rubric and to their own internalised representation of the
scoring criteria (which are affected by many
factors \cite{bejarRaterCognitionImplications2012}
such as the type of rubric and their level of
expertise) \cite{CognitiveDifferencesProficient},
and then articulating their decision through justified feedback and
grades. Assuming both the expert grader and LLM are able to parse the
response equally well, the main differences in processing lie in the
understanding and the application of rubric
contents \cite{EvaluatingQualityAI}.
Rubrics that have a discrete answer which does not permit a wide
variation of responses are most likely to be graded similarly by both
LLM and experts, as will rubrics that rely on the sort of general
knowledge that is likely to be already embedded in the LLM data set. An
example can be seen in Impey et al
(2004) \cite{impeyUsingLargeLanguage2024}
who described AI grading of short answers given by high school students
to questions in an astrobiology course, with questions such as ``What
are the advantages of large telescopes? Provide at least one''. Since
LLMs are trained on a wide variety of data ranging from web pages, books
to scientific articles and codebases like
github \cite{zhaoSurveyLargeLanguage2024},
the lack of specific training in astrobiology is not a barrier to
processing such answers to arrive at grades and feedback which are
similar to those produced by human instructors.

Poor agreement between LLM and the expert can be attributed to several
possible factors. Postgraduate medical education extends into narrow
lanes of interest that are unlikely to be featured in LLM data sets,
except on a very superficial level. The expert grader may consider other
aspects of the SAQ, for example the clinical scenario, underlying
pathology, patient-specific details given in the stem, and apply their
bedside experience to the grading. LLMs may not align with the implicit
biases and priorities of human graders. Experts might award points for
reasoning even if the final answer is incompletely correct, or permit a
wider diversity of correct approaches, recognising a plurality of
opinions in an area where there is controversy or equipoise. Subtle
differences in terminology (e.g., "PEEP titration" vs. "PEEP
optimization") may lead to different LLM responses and incorrect
grading.

The degree of agreement between LLM and the human expert in this study
has promise, considering that this study used unmodified rubrics
intended for human graders in the prompt. LLM-oriented rubric design and
careful prompt engineering is likely to produce better results. An
important property of LLMs is ``in-context learning'', the ability of
the model to learn from few examples in
context \cite{doewesLimitationsHumanComputerAgreement2021}.
Humans can perform new tasks at reasonable levels of competency given a
brief directive in natural language, and a few
examples \cite{brownLanguageModelsAre2020}.
In-context learning emulates this by using analogies and not requiring
trained datasets. Since in-context learning is prompt-based and requires
smaller data sets (``few-shot
learning'') \cite{ReasoningLargeLanguage2025},
prompt modifications have the capacity to improve performance even at
the consumer level of access. Modifications that improve outputs in
in-context learning include inducing a chain of thought, which lets the
language model break down the task at hand into smaller
steps \cite{weiChainofThoughtPromptingElicits2023},
``role playing'' where the prompt includes instructions for the LLM to
act as a particular
role \cite{gaoHumanlikeSummarizationEvaluation2023},
offering more examples (performance improves even after only five
examples) and using examples that are in a similar semantic
space \cite{liuWhatMakesGood2022}.
However, some have argued that prompting on its own cannot produce
sufficient accuracy in high-stakes
applications \cite{kapoorLargeLanguageModels2024}.
Moving forward from consumer access options, LLM prediction can be
improved using fine-tuning on a small dataset of correct and incorrect
answers \cite{kapoorLargeLanguageModels2024}.
While the training of LLMs remains out of reach for consumers, the
effectiveness of this strategy in this particular use-case remains to be
assessed.

This study has several limitations. ChatGPT was used in its default
configuration with a standardized grading prompt, which may not fully
leverage advanced prompt engineering techniques or in-context learning
strategies that could improve performance. The model relied solely on
the set of generic training data and did not have access to specialized
ICU-related literature or guidelines. The model was also evaluated using
a single domain of postgraduate medical education, specifically
mechanical ventilation, which may not generalize to other areas of
medical education where assessment styles and content complexity differ.
The sample size for this study was determined by convenience; sample
size calculation of future studies may be better informed by results of
similar research as they become available.

\section{Conclusion}
LLMs available at the consumer level differ significantly from human
graders when faced with the task of evaluating natural language answers
at a postgraduate level of study. High-stakes postgraduate assessment
demands a better standard of agreement between raters. The promising
capacity of LLM grading for certain rubric types is encouraging and the
question of whether agreement can be improved through prompt engineering
and rubric design remains unanswered. Future modifications to these
strategies may bring LLM-mediated grading of complex expert subjects
within the reach of clinician educators.

\clearpage
\bibliographystyle{ieeetr}
\bibliography{template.bib}  






\end{document}